\pdfoutput=1
\documentclass[a4paper,twoside]{article}

\usepackage{graphicx}
\DeclareGraphicsExtensions{.pdf,.png,.jpg}
\usepackage{subcaption}
\usepackage{calc}
\usepackage{amssymb}
\usepackage{amstext}
\usepackage{amsmath}
\usepackage{amsthm}
\usepackage{multicol}
\usepackage{pslatex}
\usepackage{apalike}
\usepackage{algorithm2e}
\usepackage[bottom]{footmisc}
\usepackage{hyperref}
\usepackage{multirow} 
\usepackage{xcolor}
\usepackage{SCITEPRESS}     

\begin{document}

\title{
\footnotesize\textit{\textcolor{gray}{This work was published in the Proceedings of VISAPP 2025 (VISIGRAPP 2025), SCITEPRESS.}}\\[1.5em]
\Large Deep Learning-Powered Visual SLAM Aimed at Assisting Visually Impaired Navigation
}

\author{\authorname{Marziyeh Bamdad\sup{1,2}\orcidAuthor{0000-0003-2837-7881}, Hans-Peter Hutter\sup{1}\orcidAuthor{0000-0002-1709-546X} and Alireza Darvishy\sup{1}\orcidAuthor{0000-0002-7402-5206}}
\affiliation{\sup{1}Institute of Computer Science, ZHAW School of Engineering, Obere Kirchgasse 2, 8401 Winterthur, Switzerland}
\affiliation{\sup{2}Department of Informatics, University of Zurich, 8050 Zurich, Switzerland}
\email{\{bamz, huhp, dvya\}@zhaw.ch}
}


\abstract{Despite advancements in SLAM technologies, robust operation under challenging condition such as low-texture, motion-blur, or challenging lighting remains an open challenge. Such conditions are common in applications such as assistive navigation for the visually impaired. These challenges undermine localization accuracy and tracking stability, reducing navigation reliability and safety. To overcome these limitations, we present SELM-SLAM3, a deep learning-enhanced visual SLAM framework that integrates SuperPoint and LightGlue for robust feature extraction and matching. We evaluated our framework using TUM RGB-D, ICL-NUIM, and TartanAir datasets, which feature diverse and challenging scenarios. SELM-SLAM3 outperforms conventional ORB-SLAM3 by an average of 87.84\% and exceeds state-of-the-art RGB-D SLAM systems by 36.77\%. Our framework demonstrates enhanced performance under challenging conditions, such as low-texture scenes and fast motion, providing a reliable platform for developing navigation aids for the visually impaired.}

\onecolumn \maketitle \normalsize \setcounter{footnote}{0} \vfill

\section{\uppercase{Introduction}}
\label{sec:introduction}
Visual simultaneous localization and mapping (visual SLAM) plays a key role in computer vision and robotics, enabling robots and autonomous vehicles to map their surroundings while tracking their position. Visual SLAM is applicable in various fields, including enhancing navigation for blind and visually impaired (BVI) individuals.

The field of SLAM for BVI navigation has undergone significant advancements, from studies leveraging well-established SLAM techniques to developing new SLAM solutions tailored to the requirements of the visually impaired \cite{bamdad2024slam}.
Some studies directly incorporated well-known SLAM frameworks, such as ORB-SLAM and RTAB-Map, without significant modifications or justification for their suitability. Several researchers have utilized spatial tracking frameworks from existing platforms such as Google ARCore SLAM and Intel RealSense SLAM. Some have developed customized solutions tailored specifically for visually impaired navigations \cite{bamdad2024slam}. 

Despite these advancements, SLAM technologies have certain limitations. The effectiveness of SLAM systems often depends on distinct environmental features that pose challenges in challenging environments \cite{zhang2017indoor}, \cite{jin2021wearable}. Moreover, frequent localization losses in areas lacking sufficient feature points, such as blank corridors or plain walls, compromise navigation reliability \cite{hou2022knowledge}.
Achieving accurate and reliable localization, especially for aiding the navigation of the visually impaired in challenging environments, remains an open problem.
Traditional SLAM algorithms often fail under these conditions. Although recent deep-learning SLAM models show promise, issues, such as computational complexity and data dependency, limit their application in assistive technologies for BVI. The integration of learning-based modules into traditional frameworks has improved efficiency but still requires refinement for diverse environments.

SLAM methods can be classified as direct, optical flow, and feature-based. Although direct methods are robust in low-texture scenes, they struggle in gradient-limited environments. Optical flow approaches handle environmental changes but are vulnerable to rapid movements or lighting fluctuations. Feature-based SLAM methodologies, on the other hand, excel by identifying and tracking unique features within an environment, thereby providing high precision and efficiency.

Building on this understanding and motivated by the goal of enhancing navigation aids for visually impaired individuals, we developed SELM-SLAM3 to address the challenges faced by such applications. It is developed based on ORB-SLAM3, a state-of-the-art feature-based visual SLAM system. It integrates SuperPoint for robust feature extraction and LightGlue for precise feature matching, enhancing feature quality and matching precision. This contributes to accurate pose estimation and improves localization and navigation reliability, particularly under texture-limited or dynamic lighting conditions. Evaluations of the TUM RGB-D \cite{sturm2012benchmark}, ICL-NUIM \cite{handa2014benchmark}, and TartanAir \cite{wang2020tartanair} datasets demonstrated SELM-SLAM3's superior tracking accuracy and robustness compared to ORB-SLAM3 and state-of-the-art systems, consistently achieving lower absolute trajectory errors (ATE). These results highlight the potential of improving navigation aids for BVI users in challenging scenarios. The SELM-SLAM3 implementation is available at \href{https://github.com/banafshebamdad/SELM-SLAM3}{https://github.com/banafshebamdad/SELM-SLAM3}. 

The key contributions of this study are as follows:
\begin{itemize}
    \item Enhanced feature detection and matching: By employing SuperPoint for feature extraction and LightGlue for feature matching, SELM-SLAM3 enhances the quantity and quality of the detected features and the precision of feature matching.
    \item Robust performance under adverse conditions: The system demonstrated superior tracking capability, particularly in challenging conditions, such as texture-poor scenarios.
    \item Adaptability and efficiency: The system's adaptability and efficiency in feature matching are significantly improved by the capability of LightGlue to dynamically adjust the matching process based on the complexity of each frame.
    \item Superior comparative performance: Compared with state-of-the-art systems, SELM-SLAM3 demonstrates superior performance in terms of matching accuracy and pose estimation, highlighting its effectiveness.
\end{itemize}

\section{\uppercase{Related Work}}
\subsection{SLAM-Based Navigation Solutions for BVI}
Visual SLAM technologies have been widely explored for the development of navigation aids for visually impaired individuals. These solutions can be categorized into three main approaches \cite{bamdad2024slam}. The first category leverages established SLAM frameworks such as ORB-SLAM in their solutions. For instance, \cite{son2022wearable} adopted ORB-SLAM2 for crosswalk navigation and \cite{plikynas2022indoor} leveraged ORB-SLAM3 for indoor navigation. However, these solutions are not specifically designed to perform effectively in demanding environments, such as low-texture areas or poor lighting, which are typical navigation challenges for the visually impaired.
 
The second category includes solutions that utilize spatial tracking frameworks from platforms such as ARCore, ZED cameras, and Apple's ARKit. For example, \cite{zhang2019arcore} took advantage of an ARCore-supported smartphone to track the pose and build a map of the surroundings in real-time. Although these platforms provide ready-to-use SLAM capabilities, they offer limited flexibility for optimization and enhancement, making it difficult to adapt them to the specific requirements of visually impaired navigation. 

The third category comprises customized SLAM solutions tailored specifically for visually impaired navigations. For example, the VSLAMMPT \cite{jin2020combining} was developed to assist visually impaired individuals in navigating indoor environments. These specialized solutions often focus on addressing the specific aspects of BVI navigation while using traditional SLAM techniques for localization and mapping, with their inherent limitations under challenging conditions.
 
\subsection{From Classical SLAM to Deep Learning Techniques}
Visual SLAM technologies have evolved significantly, transitioning from traditional algorithms such as ORB-SLAM3 \cite{campos2021orb}, LSD-SLAM \cite{engel2015large}, DSV-SLAM \cite{mo2021fast}, VINS-mono \cite{qin2018vins}, and OpenVSLAM \cite{sumikura2019openvslam}, which rely on handcrafted features and matching techniques to incorporate advanced deep learning-based approaches. Traditional SLAM systems have been foundational, but they often encounter issues in challenging environments.

Recently, deep learning models have introduced enhanced robustness and accuracy in SLAM applications. End-to-end deep learning visual odometry and SLAM systems, such as DPVO \cite{teed2024deep}, DROID-SLAM \cite{teed2021droid}, iDF-SLAM \cite{ming2022idf}, TrianFlow \cite{zhao2020towards}, DeepVO \cite{wang2017deepvo}, and TartanVO \cite{wang2021tartanvo} have demonstrated potential for overcoming traditional limitations. However, these systems occasionally encounter challenges. Issues such as complexity, high computational demands, and data dependency can limit their effectiveness. This becomes especially problematic for applications like navigation aids for the visually impaired, where practical constraints on sensor carriage and computational resources exist.

Another new trend in visual SLAM is the integration of learning-based modules with the traditional SLAM frameworks. Systems such as SuperPointVO \cite{han2020superpointvo}, SuperPoint-SLAM \cite{deng2019comparative} and SuperSLAM3 \cite{mollica2023integrating} have adopted deep learning for feature extraction, while retaining classical feature matching. However, the results showed that traditional matching approaches do not effectively match the learning-based feature descriptors provided by the deep learning-based feature extraction models. \cite{fujimoto2023deep} and \cite{zhu2023novel} went one step further in this area by incorporating deep-learning-based feature extraction and matching into traditional SLAM frameworks. They incorporated SuperPoint and SuperGlue as the feature extraction and feature matching modules, respectively. However, the adaptability of SuperGlue to varying conditions leaves room for further optimization. 

In this study, we introduced a novel approach by integrating SuperPoint and LightGlue into the front end of the ORB-SLAM3 framework. LightGlue offers several enhancements over SuperGlue, a previously established state-of-the-art method for sparse matching. It is optimized for both memory and computation, which are crucial for the real-time processing requirements of SLAM systems. Moreover, LightGlue's ability to adapt to the difficulty of matching tasks and provide faster inference for intuitively easy-to-match image pairs makes it suitable for latency-sensitive applications such as SLAM. 

\section{\uppercase{Implementation}}
\label{sec:implementation}
\begin{figure*}[!h]
    \centering
    \includegraphics[width=0.85\textwidth]{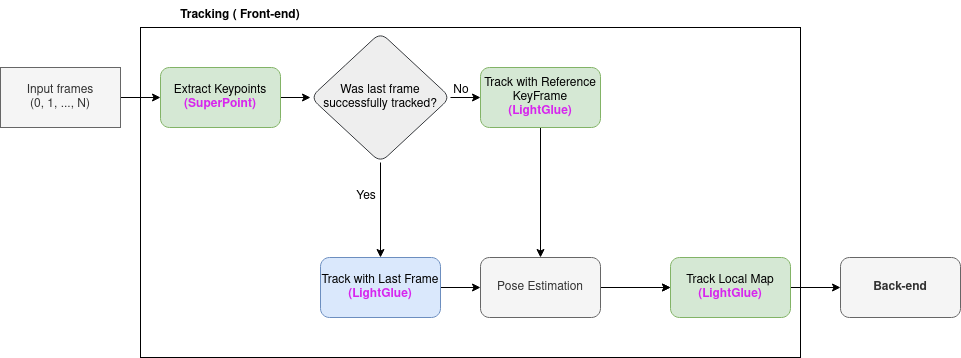}
    \caption{SELM-SLAM3 front-end, highlighting the integration of SuperPoint and LightGlue. The green boxes indicate the modified modules in ORB-SLAM3; the blue box represents the new module implemented in SELM-SLAM3; and the gray boxes depict the original modules of ORB-SLAM3 without modifications.}
    \label{fig:SELM-SLAM3_frontend}
\end{figure*}

 The main objective of SELM-SLAM3 is to enhance the feature detection and matching accuracy under challenging conditions, such as changing lighting, motion blur, and poor textures. As illustrated in Figure \ref{fig:SELM-SLAM3_frontend}, the system builds on ORB-SLAM3, a state-of-the-art SLAM framework known for its accuracy and processing speed, with modifications focused on its front end. The deep learning-based SuperPoint model is used to identify and describe keypoints, ensuring the reliable detection of high-quality features in diverse environments. LightGlue complements this by accurately matching the features between consecutive frames and keyframes, which are crucial for consistent and precise tracking.

Integrating SuperPoint and LightGlue into ORB-SLAM3 requires addressing the differences in data formats, structures, and methodologies. These adjustments enabled the seamless incorporation of advanced feature detection and matching algorithms into the SLAM framework. Essential libraries, including OpenCV, were employed to manage the keypoints, descriptors, feature normalization, and data access. The pre-trained SuperPoint and LightGlue models, provided in the ONNX format, were executed on a GPU for efficiency, while the other system components ran on a CPU. These models, sourced from the AIDajiangtang GitHub repository \cite{dajiangtang_github}, delivered satisfactory test performance, eliminating the need for fine-tuning at this stage.


\subsection{Feature Extraction}
 This study demonstrates the transition from traditional feature extraction methods to enhanced deep learning models in visual SLAM systems. ORB-SLAM3 uses a multiscale strategy to extract features and detect FAST corners across eight scale levels with a scale factor of 1.2. To ensure uniform feature distribution, it subdivides each scale level into grids, identifying a minimum of five corners per grid cell, and dynamically adjusting thresholds to detect adequate corners, even in low-texture areas. The orientation and ORB descriptors are then computed for the selected corners, as detailed in \cite{mur2015orb}. In contrast, SuperPoint extracts features from full-resolution images, capturing a broader array of features while preserving the fine details missed by ORB-SLAM3’s pyramid approach. Trained by scale invariance and pattern recognition, SuperPoint addresses the limitations of traditional handcrafted methods. SELM-SLAM3 incorporates SuperPoint for feature extraction, transforms each frame into grayscale, and employs the SuperPoint detector to identify key features, which are then utilized throughout the SLAM pipeline. By leveraging SuperPoint’s ability to recognize intricate patterns and their scale invariance, SELM-SLAM3 captures high-quality features, even under challenging conditions such as low textures or poor lighting.

\subsection{Feature Matching}
SELM-SLAM3 employs the LightGlue model to match features between frames. LightGlue is an advanced neural network designed to enhance the sparse local feature matching across images. Building on the achievements of its predecessor, SuperGlue, LightGlue combines the power of attention and transformer mechanisms into the matching problem. 
LightGlue can dynamically adjust its computational intensity based on the complexity of each image pair owing to its ability to assess the prediction confidence.
This adaptability allows LightGlue to surpass SuperGlue in terms of speed, precision, and training ease, making it an excellent choice for various computer vision applications \cite{lindenberger2023lightglue}.
The substitution of ORB-SLAM3 traditional matching algorithms with LightGlue necessitates certain adjustments to the SLAM pipeline. 
The guided matching approach, which involves projecting features and searching for matches within a constrained window, was eliminated to fully leverage the capabilities of LightGlue. Similarly, computationally intensive Bag of Words (BoW)-based matching was replaced, to enhance the feature matching process.

\subsection{Tracking}
 Tracking is responsible for localizing the camera and deciding when to insert a new keyframe, and relies heavily on effective feature extraction and matching. Robust feature extraction ensures that sufficient keypoints are available for tracking, even in challenging scenarios. Efficient feature matching associates keypoints across frames, enabling accurate camera pose estimation. 
 
 In ORB-SLAM3, tracking follows a constant-velocity motion model to predict the camera pose using a guided search for the map points observed in the last frame. Features are projected onto the current frame using their 3D position and initial pose estimation with matches searched within a small window. However, inaccuracies in the motion model can result in insufficient matches, requiring the system to compute bag-of-words (BoW) vectors for the current and reference keyframes, although this approach is computationally demanding. 
 
 SELM-SLAM3 addresses these limitations by bypassing 3D point projection, focusing on direct feature matching between the previous and current frames, increasing the number of matches, and improving tracking accuracy. Regardless of the initial method, the Track Local Map module is applied to secure more matches. ORB-SLAM3 uses reprojection to match local map points, whereas SELM-SLAM3 uses LightGlue to match these points directly with the current frame. Only map points within the field of view of the current frame were considered, and their projected coordinates were treated as inputs to LightGlue. 
 
 The final tracking step involves deciding whether to insert a new keyframe based on criteria such as the number of frames since the last keyframe, and whether the current frame tracks less than 90\% of its reference keyframe. SELM-SLAM3 inserts fewer keyframes than ORB-SLAM3, while providing a denser map because of its ability to extract more features per frame and achieve more accurate matches. These advancements from integrating SuperPoint and LightGlue significantly enhanced the tracking performance and map density in SELM-SLAM3.

\section{\uppercase{Experements and Results}}
\label{sec:experiments}
We evaluated SELM-SLAM3 on TUM RGB-D \cite{sturm2012benchmark} freiburg1 sequences, ICL-NUIM \cite{handa2014benchmark}, and TartanAir \cite{wang2020tartanair} datasets by running both the baseline ORB-SLAM3 and the enhanced SELM-SLAM3 systems on selected sequences. The absolute trajectory error (ATE) was calculated by evaluating the Euclidean distances between the estimated camera poses and the ground truth \cite{sturm2012benchmark}. Tests were conducted on a system with an Intel i9-12950HX CPU, 64 GB RAM, and an NVIDIA RTX A2000 8 GB GPU running Ubuntu 20.04 LTS.
We also compared our results with those reported by \cite{fujimoto2023deep}, \cite{li2021rgb}, \cite{whelan2015real}, selected for their shared use of RGB-D sensors, testing on similar datasets, and focus on improving tracking and mapping accuracy, aligning with the primary goal of SELM-SLAM3.

\subsection{TUM RGB-D dataset evaluation}
\begin{figure}[!h]
    \centering
    \includegraphics[width=0.5\textwidth]{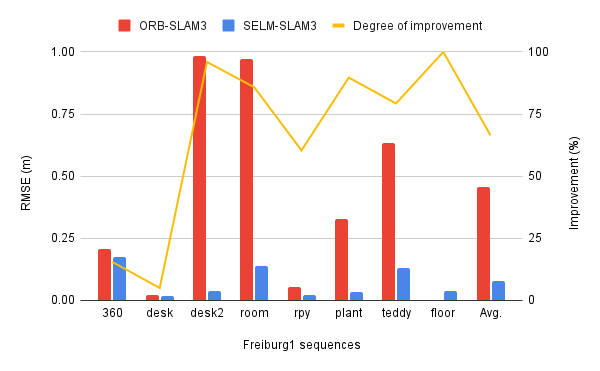}
    \caption{Comparison of ATE (m) on TUM RGB-D sequences illustrates the enhanced accuracy achieved by SELM-SLAM3 and the improvement over ORB-SLAM3.}
    \label{fig:evaluation_on_TUM}
\end{figure}

\begin{figure}[!h]
\centering
\begin{subfigure}{.22\textwidth}
  \centering
  \includegraphics[width=\linewidth]{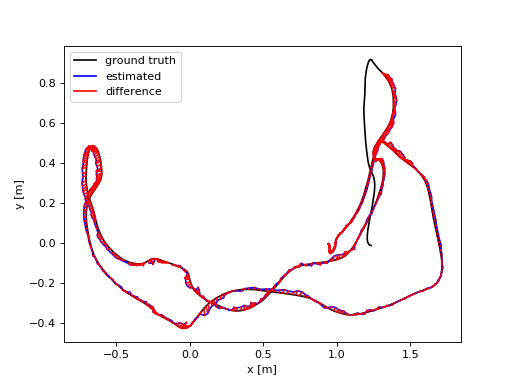}
  \caption{TUM f1 desk}
  \label{fig:TUM_desk}
\end{subfigure}
\begin{subfigure}{.22\textwidth}
  \centering
  \includegraphics[width=\linewidth]{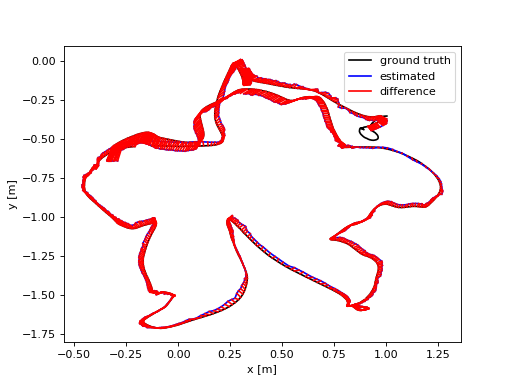}
  \caption{TUM f1 plant}
  \label{fig:f1_plant}
\end{subfigure}
\begin{subfigure}{.22\textwidth}
  \centering
  \includegraphics[width=\linewidth]{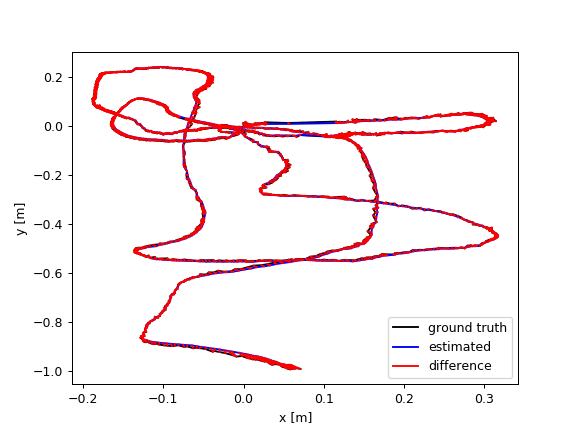}
  \caption{ICL-NUIM lr-kt0}
  \label{fig:ICL_lrkt0}
\end{subfigure}
\begin{subfigure}{.22\textwidth}
  \centering
  \includegraphics[width=\linewidth]{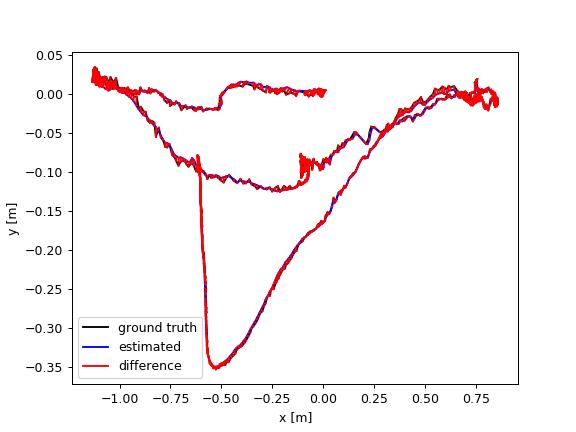}
  \caption{ICL-NUIM or-kt2}
  \label{fig:of_kt2}
\end{subfigure}
\caption{Sample of estimated trajectory and ground-truth.}
\label{fig:estimated_trajectories}
\end{figure}

The TUM RGB-D dataset includes color and depth images from diverse indoor environments, captured at 30 Hz with a resolution of 640 × 480. ORB-SLAM3 was configured with eight scale levels, a scale factor of 1.2, and 1000 features. For both ORB-SLAM3 and SELM-SLAM3, the loop-closing module was disabled to ensure a fair and focused comparison of the core odometry capabilities with the results reported in \cite{fujimoto2023deep}, which describes an RGB-D odometry system. Both systems were evaluated on selected sequences and the absolute trajectory error (ATE) was calculated using the TUM RGB-D online tool. Figure \ref{fig:evaluation_on_TUM} shows a significant improvement in the ATE for SELM-SLAM3 compared to ORB-SLAM3. ORB-SLAM3 failed on the "floor" sequence, highlighting SELM-SLAM3's robustness. Table \ref{TUM_ATE_table} demonstrates consistent RMSE improvements, with SELM-SLAM3 outperforming ORB-SLAM3 across the sequences. Sample trajectories for the Freiburg1 desk and plant sequences are shown in Figures \ref{fig:TUM_desk} and \ref{fig:f1_plant}.  

\begin{table*}[h]
    \caption{Comparison of absolute trajectory error (ATE) in meters (m) on the TUM RGB-D dataset against baseline}
    \label{TUM_ATE_table}
    \centering
    \begin{tabular}{|c|c|c|c|c|c|c|c|c|c|}
        \hline
        \multirow{2}{*}{Freiburg1} & \multicolumn{4}{c|}{ORB-SLAM3} & \multicolumn{4}{c|}{SELM-SLAM3} & \multirow{2}{*}{RMSE \% Boost} \\
        \cline{2-5} \cline{6-9}
        & RMSE & Mean & Median & S.D. & RMSE & Mean & Median & S.D. & \\
        \hline
        360 & 0.207 & 0.2 & 0.201 & 0.054 & 0.175 & 0.16 & 0.147 & 0.071 & 15.46 \\ \hline
        desk & 0.02 & 0.017 & 0.016 & 0.009 & 0.019 & 0.017 & 0.014 & 0.01 & 5 \\ \hline
        desk2 & 0.982 & 0.948 & 0.927 & 0.256 & 0.04 & 0.035 & 0.033 & 0.019 & 95.93 \\ \hline
        room & 0.971 & 0.913 & 0.929 & 0.329 & 0.138 & 0.129 & 0.119 & 0.05 & 85.79 \\ \hline
        rpy & 0.053 & 0.047 & 0.045 & 0.023 & 0.021 & 0.017 & 0.015 & 0.011 & 60.38 \\ \hline
        plant & 0.329 & 0.243 & 0.171 & 0.221 & 0.034 & 0.03 & 0.025 & 0.015 & 89.67 \\ \hline
        teddy & 0.632 & 0.528 & 0.355 & 0.347 & 0.131 & 0.104 & 0.076 & 0.079 & 79.27 \\ \hline
        floor & x & x & x & x &  0.04 & 0.034 & 0.026 & 0.022 & 100 \\
        \hline
        Avg. & 0.456 & 0.414 & 0.378 & 0.177 & 0.08 & 0.07 & 0.061 & 0.036 & 66.44 \\
        \hline
    \end{tabular}
\end{table*}

We also compared our system's results with \cite{fujimoto2023deep} and \cite{whelan2015real}. The method in \cite{fujimoto2023deep} employs SuperPoint for feature extraction and SuperGlue for feature matching, enhancing the accuracy and robustness of RGB-D odometry in challenging environments. \cite{whelan2015real} uses RGB-D sensors for real-time surface reconstructions, leveraging volumetric reconstruction and a GPU-based 3D cyclical buffer for unbounded space mapping. This method improves pose estimation with dense geometric and photometric constraints, and features efficient map updates with space deformation for loop closure. Table \ref{tab:TUM_ATE_comparison_table} highlights SELM-SLAM3’s superior performance across most of the tested Freiburg1 sequences in the TUM RGB-D dataset. SELM-SLAM3 achieved the lowest ATE in all sequences compared with the deep-feature-based approach in \cite{fujimoto2023deep} and outperformed \cite{whelan2015real} in six out of seven sequences.
\begin{table}[h]
    \caption{Comparison of ATE (m) on TUM RGB-D dataset. DL-based: Deep Learning-based \cite{fujimoto2023deep}; Volumetric: Volumetric-based \cite{whelan2015real}. NA indicates that ATE values were not reported.}
    \label{tab:TUM_ATE_comparison_table}
    \centering
    \begin{tabular}{|c|c|c|c|}
        \hline
        Seq. & SELM-SLAM3 & DL-based & Volumetric \\ \hline
        desk  & 0.019 & 0.091 & 0.040 \\ \hline
        desk2 & 0.04 & 0.087 & 0.074 \\ \hline
        room & 0.138 & 0.327 & 0.081 \\ \hline
        rpy & 0.021 & NA & 0.031 \\ \hline
        plant & 0.034 & 0.076 & 0.050 \\ \hline
        teddy & 0.131 & 0.555 & NA \\ \hline
        floor & 0.04 & 0.348 & NA \\
        \hline
    \end{tabular}
\end{table}

\subsection{ICL-NUIM dataset evaluation}
\begin{figure}[!h]
    \centering
    \includegraphics[width=0.5\textwidth]{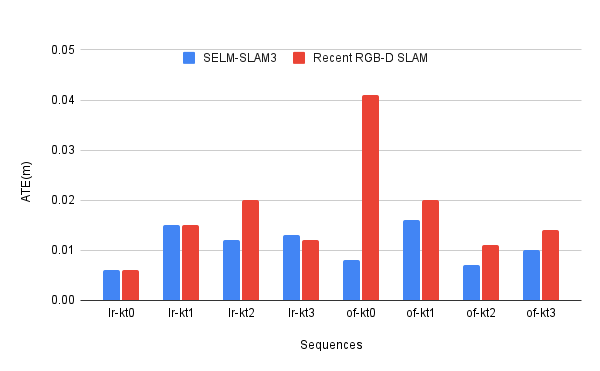}
    \caption{Comparison of ATE on ICL-NUIM dataset.}
    \label{fig:evaluation_on_ICL_NUIM}
\end{figure}

\begin{figure}[!h]
\centering
\begin{subfigure}{.15\textwidth}
  \centering
  \includegraphics[width=\linewidth]{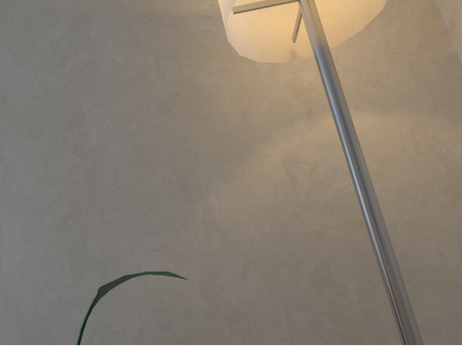}
  \caption{Input}
  \label{fig:input_image}
\end{subfigure}
\begin{subfigure}{.15\textwidth}
  \centering
  \includegraphics[width=\linewidth]{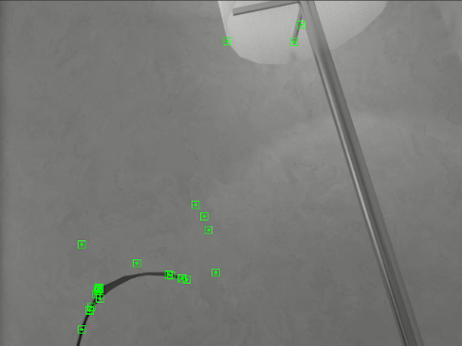}
  \caption{ORB-SLAM3}
  \label{fig:orb_slam3}
\end{subfigure}
\begin{subfigure}{.15\textwidth}
  \centering
  \includegraphics[width=\linewidth]{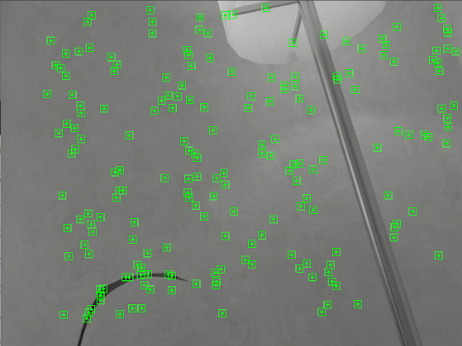}
  \caption{SELM}
  \label{fig:selm_slam3}
\end{subfigure}
\caption{Challenging scene in lr-kt3}
\label{fig:texture_poor}
\end{figure}

The ICL-NUIM dataset was used to benchmark the RGB-D visual odometry and SLAM algorithms. Its synthetic scenes and extensive low-textured surfaces, such as walls, ceilings, and floors, pose specific challenges for tracking and mapping owing to the lack of distinct features. The dataset includes living room sequences for benchmarking camera trajectory and reconstruction (with 3D ground truth, depth maps, and camera poses), and office room sequences for camera trajectory benchmarking. We tested SELM-SLAM3 across all sequences of the ICL-NUIM dataset and compared the results with \cite{li2021rgb}, which introduced a robust RGB-D SLAM system tailored for structured environments. This system enhances the tracking and mapping accuracy by leveraging geometric features such as points, lines, and planes, employing a decoupling-refinement method for pose estimation with Manhattan World assumptions, and using an instance-wise meshing strategy for dense map construction. The results, shown in Figure \ref{fig:evaluation_on_ICL_NUIM}, and Table \ref{tab:ICL_NUIM_ATE_table}, demonstrate that SELM-SLAM3 outperformed both ORB-SLAM3 and the RGB-D SLAM system from \cite{li2021rgb}. Figure \ref{fig:texture_poor} further highlights SELM-SLAM3’s superior tracking capability, successfully handled texture-poor scenarios in which ORB-SLAM3 experienced tracking loss. Sample trajectories are shown in Figures \ref{fig:ICL_lrkt0} and \ref{fig:of_kt2}.

\begin{table}[h]
    \caption{Comparison of ATE (m) on ICL-NUIM dataset. SELM: SELM-SLAM3; Recent SLAM \cite{li2021rgb}.}
    \label{tab:ICL_NUIM_ATE_table}
    \centering
    \begin{tabular}{|c|c|c|c|}
        \hline
        Seq. & SELM & Recent SLAM & ORB-SLAM3 \\ \hline
        lr-kt0  & 0.006 & 0.006 & 0.603 \\ \hline
        lr-kt1 & 0.015 & 0.015 & 0.293 \\ \hline
        lr-kt2 & 0.012 & 0.02 & 0.826 \\ \hline
        lr-kt3 & 0.013 & 0.012 & 1.12 \\ \hline
        of-kt0 & 0.008 & 0.041 & 0.639 \\ \hline
        of-kt1 & 0.016 & 0.02 & 0.964 \\ \hline
        of-kt2 & 0.007 & 0.011 & 0.768 \\ \hline
        of-kt3 & 0.01 & 0.014 & 0.757 \\ \hline
        Avg. & 0.0108 & 0.0173 & 0.746 \\ \hline
    \end{tabular}
\end{table}

\subsection{TartanAir dataset evaluation}
The TartanAir dataset, with its extensive size and diversity, is a challenging benchmark comprising 1037 sequences, each with 500-4000 frames, totaling over one million frames. Collected in photorealistic simulation environments with realistic lighting, it includes diverse motion patterns (e.g., hands, cars, robots, and MAVs) across 30 environments spanning urban, rural, natural, domestic, public, and science-fiction settings. TartanAir categorizes sequences into Easy, Medium, and Hard difficulty levels to evaluate SLAM systems under varying complexities.  

\begin{figure}[!h]
    \centering
    \includegraphics[width=0.40\textwidth]{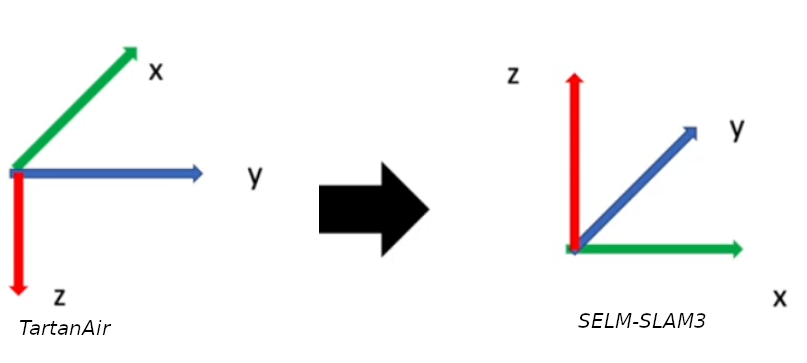}
    \caption{Transformation between NED frame of TartanAir and pose coordinate system in SELM-SLAM3.}
    \label{fig:frame_transformation}
\end{figure}

We focused on the "Hospital" sequence at the "Hard" difficulty level, which is relevant to visually impaired navigation. A specialized tool was developed to convert TartanAir data into a format compatible with SELM-SLAM3 and ORB-SLAM3, allowing custom dataset sequences. Additionally, the depth information was modified to meet the system requirements, and trajectory transformations aligned the North-East-Down (NED) frame of TartanAir with the reference frames of SELM-SLAM3 and ORB-SLAM3 (Figure \ref{fig:frame_transformation}). This ensured compatibility of the pose data with initial poses adjusted from TartanAir’s coordinates (7, -30, 3) to (0, 0, 0).  

\begin{figure}[!h]
    \centering
    \includegraphics[width=0.5\textwidth]{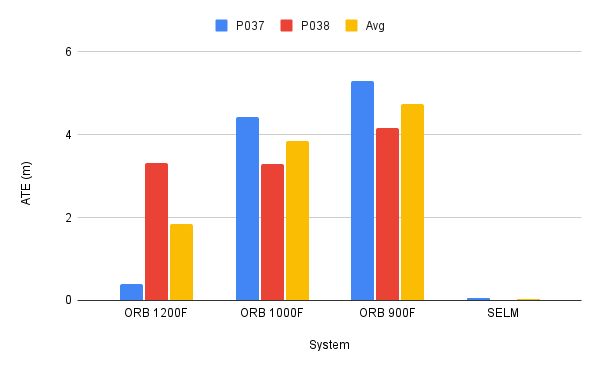}
    \caption{ATE on the TartanAir Hospital sequences.}
    \label{fig:TartanAir_ATE}
\end{figure}

The evaluation involved testing both systems on selected sequences with ORB-SLAM3 configured using eight scale levels, a scale factor of 1.2, and feature point settings of 750, 900, and 1000. The results for the "Hospital-Hard" sequences are shown in Figure \ref{fig:TartanAir_ATE} and Table \ref{tab:TartanAir_ATE_table}. SELM-SLAM3 exhibited significantly lower ATE values, demonstrating its efficacy in the challenging TartanAir environment.  

\begin{table}[h]
    \caption{Comparison of ATE (m) on TartanAir dataset.}
    \label{tab:TartanAir_ATE_table}
    \centering
    \begin{tabular}{|c|c|c|c|}
        \hline
        Systems & P037 & P038 & Avg. \\ \hline
        ORB-SLAM3 1200F  & 0.389 & 3.308 & 1.848 \\ \hline
        ORB-SLAM3 1000F &  4.428 & 3.294 & 3.861 \\ \hline
        ORB-SLAM3 900F & 5.298 & 4.167 & 4.733 \\ \hline
        SELM-SLAM3 & 0.049 & 0.020 & 0.035 \\ \hline
    \end{tabular}
\end{table}

SELM-SLAM3’s advanced feature extraction and matching yielded more accurate and abundant features, enhancing the tracking accuracy with ATE below 2 cm for ICL-NUIM, below 4 cm for TartanAir, and below 8 cm for TUM RGB-D. In contrast, ORB-SLAM3 showed a significantly lower accuracy, with instances of tracking loss and complete failure in one sequence.

\section{\uppercase{Conclusions}}
\label{sec:conclusion}
This study introduced SELM-SLAM3, an enhanced visual SLAM framework, as a foundational step toward solutions for visually impaired navigation. By incorporating advanced deep-learning-based techniques, SuperPoint for feature extraction and LightGlue for feature matching, SELM-SLAM3 significantly improves pose estimation and tracking accuracy. SELM-SLAM3 outperformed ORB-SLAM3 with an average improvement of 87.84\% and surpassed state-of-the-art RGB-D SLAM systems with an average enhancement of 36.77\% in pose estimation accuracy, particularly under challenging conditions, such as low texture, rapid motion, and changing lighting. Comprehensive evaluations of the TUM RGB-D, ICL-NUIM, and TartanAir datasets confirmed their effectiveness. Its adaptability to diverse environments underscores SELM-SLAM3's potential for real-world applications, particularly in aiding BVI navigation, where reliability and precision are critical. However, current evaluations rely on standard datasets, which may not fully reflect the real-world complexities. Future studies should test the system on tailored datasets to ensure broader applicability. Future work could also focus on optimizing computational performance to ensure real-time applicability.

\bibliographystyle{apalike}
{\small
\bibliography{main}}

\end{document}